\documentclass[lettersize,journal]{IEEEtran}

\usepackage{times}
\usepackage{latexsym}
\usepackage{amsmath}
\usepackage{changepage}
\usepackage{float}
\usepackage{array}
\usepackage{graphicx}
\usepackage{amssymb}
\usepackage{algorithmic}
\usepackage{tabularx}
\usepackage{textcomp}
\usepackage{caption}
\usepackage{subcaption} 
\usepackage{xcolor}
\usepackage{enumitem}
\usepackage{multirow}
\usepackage{ragged2e}

\definecolor{normalGreen}{RGB}{141, 208, 63}

\usepackage{booktabs, threeparttable, pifont}
\newcommand{\cmark}{\ding{51}}   
\newcommand{\xmark}{\ding{55}}   

\usepackage[T1]{fontenc}

\usepackage[utf8]{inputenc}

\usepackage{microtype}

\usepackage{inconsolata}

\usepackage[colorlinks,urlcolor=blue,linkcolor=blue,citecolor=blue]{hyperref}

\title{DTO: a Differentiable Training Objective\\for Effective Counterfactual Story Rewriting}

\author{Amelie~Girard~and~Massimo~Piccardi,~\IEEEmembership{Senior~Member,~IEEE}
\thanks{This manuscript was submitted for review the X of March 2026.}
\thanks{A. Girard is with University of Technology Sydney, Sydney, NSW 2007, Australia (e-mail: amelie.girard@uts.edu.au).}
\thanks{M. Piccardi is with University of Technology Sydney, Sydney, NSW 2007, Australia (e-mail: massimo.piccardi@uts.edu.au).}
}

\begin{document}

\maketitle

\begin{abstract}
Counterfactual story rewriting is a natural language processing task that requires updating an existing story to reflect a chosen alternative event, yet preserving all the unaffected storyline elements and overall coherence. While large language models have recently made remarkable progress on this task, it still remains challenging since the required modifications are typically very small in size and highly localized. As a consequence, models trained in a conventional manner with the maximum-likelihood training objective tend to overlook these nuances. At the same time, more sophisticated training approaches based on reinforcement learning are notoriously slow and difficult to set up. For these reasons, our paper proposes a novel, differentiable training objective (DTO) that directly optimizes for the requisite counterfactual improvements. In our approach, a transformer model is fine-tuned via end-to-end backpropagation against a fully differentiable loss function that jointly rewards (i) fidelity to the reference rewrite and (ii) semantic consistency with the source narrative. The empirical evaluation on the \textsc{TimeTravel} and ART datasets shows that the proposed DTO approach has been able to surpass a maximum-likelihood baseline and a preference-based approach, and perform competitively against two contemporary large language models in all evaluation metrics. These findings substantiate the effectiveness of task-specific differentiable objectives for nuanced, controlled text-generation tasks.
\end{abstract}



\section{Introduction}
Reasoning underlies problem‑solving \cite{angeles1981dictionary}, decision‑making \cite{finocchiaro1984informal}, persuasion \cite{govier1989critical}, and explanation \cite{kahneman2011thinking}, and has been analyzed in psychology \cite{wason1972psychology}, philosophy \cite{passmore1961philosophical}, and computer science \cite{pearl2018bookofwhy}. In the natural language processing field, this capability is often formalized as natural language reasoning \cite{jin2024large}. Despite the remarkable progress of the recent years, it remains one of the most challenging frontier for the field \cite{hobbhahn2022,kosoy2022}. 

Among reasoning tasks, counterfactual story rewriting aims to probe the ability of a model to provide counterfactual modifications to a story ending \cite{lin-etal-2020-commongen}. In this task, given a story and a hypothetical condition that contradicts the initial event, a model must minimally alter the story's ending to reflect the consequences of the counterfactual event. For example, if a story originally describes a character going to a park, a counterfactual condition might specify ``what if the character stayed at home instead,'' and the model must alter the storyline accordingly with minimal edits. This task was formalized by the \textsc{TimeTravel} dataset, which provides nearly 28k instances of human-written counterfactual story revisions~\cite{qin2019counterfactual}, testing a model’s ability to maintain the narrative’s core while adapting to a counterfactual change.


A primary challenge in counterfactual rewriting lies in precisely controlling edits to preserve original content where possible, while effectively enforcing the new hypothetical condition. Early sequence-to-sequence models tended to either copy original endings with no or minimal changes, failing to properly integrate the counterfactual condition, or make excessive edits, compromising narrative coherence~\cite{chen2022educat}. To mitigate these issues, structured rewriting techniques have been proposed. For instance, \cite{hao2021sketch} proposed an unsupervised editing method leveraging causal inference to identify and selectively rewrite only causally dependent segments, further demonstrating improvements through custom metrics for minimality versus coherence trade-off.
Recently, \cite{Mu2024causal} integrated variational autoencoders (VAEs) and causality classifiers with external commonsense knowledge to further improve causal coherence and reduce undesired edits. In turn, \cite{Wang2024SCM} advanced this line by employing structural causal models (SCMs) with hierarchical latent disentanglement, allowing more precise minimal editing at multiple narrative levels.

In this paper, we introduce a different solution: a differentiable training objective (DTO) for counterfactual story rewriting. Rather than using discrete rewards or ad-hoc editing procedures, we directly embed the goals of minimal editing and consistency into the model’s training loss. We achieve this by leveraging BARTScore, a text generation evaluation metric based on a pre-trained BART model, as a differentiable reward~\cite{bartscore}. As a metric, BARTScore provides a learned measure of how well one piece of text (e.g., a rewritten story) reflects the content of another text (e.g., the original story and a counterfactual event). By employing BARTScore as our training objective, we are able to reward the model for producing outputs that remain faithful to the original narrative, 
while still being valid given the counterfactual premise.
Since BARTScore itself is computed via a neural model, we are able to backpropagate through this metric to the generative model by ``softening'' the generation of the rewritten ending, i.e. generating expected embeddings rather than discrete tokens. This approach effectively treats an evaluation metric as part of the training signal, aligning the model’s gradients with the end task of accurate yet minimal rewriting.

Our proposed DTO approach follows a line of research in direct loss shaping \cite{wang2019debleu, jauregi2021berttune, liu2022eisl} that attempts to avoid the pitfalls of non-differentiable policy-gradient approaches such as REINFORCE \cite{williams1992simple}. Following this line of research, we apply a differentiable objective to the narrative rewriting domain for the first time: an approach that uses the feedback from BARTScore to guide learning, yielding a more stable training process (no sampling-induced noise) and tightly coupling the training signal to an evaluation criterion for the task.

Our experiments on the \textsc{TimeTravel} benchmark show that our DTO-based training strategy achieves superior performance in maintaining narrative fidelity and ensuring accurate rewrites compared to traditional supervised learning methods. Specifically, our method generates stories closer to human-authored counterfactual endings with fewer extraneous edits, as reflected by improved metrics of content preservation. Our key contributions can be summarized as:

\begin{enumerate}
    
    \item We propose a novel training approach, DTO, which explicitly optimizes a generative model with respect to counterfactual rewriting metrics. 
    
    \item We evaluate different reward functions based on BARTScore \cite{bartscore} that explore different trade-offs between adherence to the expected ending and differentiation from the original ending.
    
    
    \item Our results over the \textsc{TimeTravel} \cite{qin2019counterfactual} dataset show that our model has been able to outperform the maximum-likelihood baseline and a preference-based model, and also perform competitively against two contemporary large language models (LLMs). An ablative and qualitative analysis have further confirmed its strong performance.  
    
\end{enumerate}

The rest of our paper is organized as follows: Section~\ref{sec:related} reviews the main literature on counterfactual text generation and on differentiable training objectives for generative tasks. Section~\ref{sec:methodology} presents our DTO-based rewriting approach, describing the loss function and the training algorithm in detail. Section~\ref{sec:experiments} describes the experimental set-up, while Section \ref{sec:results} presents the results and the discussion. Finally, Section~\ref{sec:Conclusion} recapitulates our findings and concludes the paper.\footnote{All our code can be downloaded from: \url{https://github.com/amelie-girard7/TimeTravel-DifferentiableMetrics}.}

\section{Related Work}
\label{sec:related}

Counterfactual rewriting as a formal NLP task began with \cite{qin2019counterfactual}, who introduced the \textsc{TimeTravel} dataset. Subsequent research has primarily focused on structured rewriting methods to control the balance between minimal edits and coherence. \cite{hao2021sketch} proposed a two-step sketch-and-customize framework, explicitly ensuring minimal yet necessary edits. \cite{chen2022educat} introduced \textsc{EduCat}, employing causal inference to identify causally invariant parts of the narrative, enabling focused edits that preserve overall coherence. More recently, \cite{Mu2024causal} and \cite{Wang2024SCM} advanced the field further by integrating causal inference and structural causal models (SCMs), respectively, to enhance causal consistency in counterfactual generation. These methods primarily rely on explicit constraints or inference-time heuristics rather than training-time optimization of rewriting objectives. Eventually, \cite{yang2024harnessing} provided a comprehensive analysis of the performance of LLMs in counterfactual reasoning tasks.

\paragraph{Evaluation metrics}

Automated evaluation metrics offer a practical alternative to human assessments thanks to their low computational cost. Table~\ref{tab:metrics} provides a concise overview of the most widely used automatic evaluation metrics for text-generation research. Early evaluation methods primarily measured lexical similarity using manually-crafted rules. Metrics such as BLEU~\cite{bleu} and ROUGE~\cite{rouge} are prominent examples, relying heavily on n-gram overlap to assess quality. However, these methods often fall short when evaluating paraphrased outputs that are semantically accurate but differ lexically, leading to misleadingly low scores~\cite{callison2006re}. 
By design, these metrics are highly sensitive to surface-level word variation and fail to adequately capture deeper semantic or syntactic differences compared to reference texts. As a result, their alignment with human judgment is often weak, especially when evaluating systems with similar performance levels~\cite{novikova2017we}. 

An early attempt to improve on this was provided by METEOR~\cite{banerjee2005meteor}, which incorporated synonym matching, token alignment, and hyperparameter training. However, a significant increase in flexibility has been achieved by more recent metrics such as COMET~\cite{comet}, BERTScore~\cite{bertscore}, and BLEURT~\cite{sellam2020bleurt}, which all utilize embeddings from pre-trained neural networks to better capture the semantic and contextual nuances of language. At its turn, BARTScore~\cite{bartscore} has framed evaluation as a text-generation task: it computes the likelihood of generating the hypothesis (or reference) using a pre-trained BART model, conditioned on either the source, reference, or both. Unlike matching-based metrics, BARTScore can directly model semantic plausibility without explicit alignment, offering flexibility across tasks like summarization and translation. However, its reliance on generative probabilities may introduce biases from the underlying model's training data.

\begin{table}[t]
  \scriptsize
  \setlength{\tabcolsep}{2pt}
  \renewcommand{\arraystretch}{0.90}
  \centering
  \begin{threeparttable}
    \begin{tabularx}{\columnwidth}{@{}l c p{0.25\columnwidth} c c c c@{}}
      \toprule
      \textbf{Metric} & \textbf{Trained} & \textbf{Paradigm} &
      \textbf{(S,H)} & \textbf{(R,H)} & \textbf{(S,R,H)} & \textbf{Task}\\
      \midrule
      ROUGE      & \xmark & Match          &  &  \cmark      &        & SUM \\
      BLEU       & \xmark & Match          &        & \cmark &        & MT  \\
      CHRF       & \xmark & Match          &        & \cmark &        & MT  \\
      METEOR       & \xmark & Match          &        & \cmark &        & MT  \\
      BERTScore  & \cmark & Match          &        & \cmark &        & MUL \\
      MoverScore & \cmark & Match          &        & \cmark &        & MUL \\
      PRISM      & \cmark & Paraphrase     & \cmark &        &        & MT  \\
      BLEURT     & \cmark & Regress        &        & \cmark &        & MT  \\
      S3         & \cmark & Regress        &        &        & \cmark & SUM \\
      VRM        & \cmark & Regress        &        & \cmark &        & SUM \\
      COMET      & \cmark & Regress/Rank   &        & \cmark &        & MT  \\
      BEER       & \cmark & Rank           &        & \cmark &        & MT  \\
      BARTScore  & \cmark & Generation     & \cmark & \cmark &        & MUL \\
      \bottomrule
    \end{tabularx}

    \caption{Widely used metrics for text generation\tnote{a}}
    \label{tab:metrics}
    \begin{tablenotes}[flushleft]
      \footnotesize
      \item[a]\textbf{Legend.}  
        $(\mathbf{S,H})$: uses \emph{source} \& \emph{hypothesis}.  
        $(\mathbf{R,H})$: uses \emph{reference} \& \emph{hypothesis}.  
        $(\mathbf{S,R,H})$: uses all three.  
        \textbf{Tasks}: \textbf{SUM} = Summarization, \textbf{MT} = Machine Translation, \textbf{MUL} = Multiple.  
        Settings/tasks included only if supported by the original paper. Source: \cite{bartscore}.
    \end{tablenotes}
  \end{threeparttable}
\end{table}

Notably, conventional metrics for natural language generation may not be able to adequately reflect the quality of a counterfactual rewriting. For this reason, \cite{chen2022educat} introduced a specialized evaluation metric that is able to better quantify the trade-off between coherence and minimal edits. This line of research highlights the central challenge of the task and the need for approaches that can balance this twofold goal.

\paragraph{Differentiable Training Objectives}

The universal training objective for natural language generation models is the negative log-likelihood (NLL -- interchangeably referred to as cross entropy), where training aims to maximize the probability of the tokens of a given reference \cite{BahdanauCB14}. This training objective is differentiable by design, in contrast, for instance, to reinforcement learning approaches that are collectively referred to as non-differentiable \cite{williams1992simple}. Several other differentiable training objectives have been proposed in recent years, most notably direct preference optimization (DPO) \cite{dpo2023} and its variants (e.g., contrastive preference optimization (CPO) \cite{Xu2024CPO}), where the model is trained by using two or more ranked references (see Appendix \ref{app:CPO} for details). However, only a few papers to date have proposed directly using an evaluation metric as a differentiable training objective. This goal may be appealing because it promises to align the model's performance with the desired evaluation measure, while at the same time avoiding the complications of non-differentiable approaches such as reinforcement learning. Among them, \cite{wang2019debleu} has reformulated BLEU as a differentiable loss to circumvent the need for policy-gradient approaches;
\cite{jauregi2021berttune} has used BERTScore to train a machine translation model;
\cite{liu2022eisl} has used the Edit-Invariant Sequence Loss (EISL) as the training objective to achieve robustness to minor paraphrasing;
and \cite{girard2026TALLIP} has proposed a simple modification to the negative log-likelihood suiting the counterfactual story rewriting task. 
These methods, that directly optimize high-level metrics, align most closely with our proposed DTO approach.

\section{Methodology}
\label{sec:methodology}

\subsection{Task Definition}
\label{subsec:task}

In the definition of the counterfactual story rewriting task provided by \cite{qin2019counterfactual}, a story is divided into five labeled elements as shown in Table~\ref{tab:sample_structure}. The first three, (i) premise ($X_P$), (ii) initial event ($X_{IE}$), and (iii) original ending ($X_{OE}$) constitute the original story. The remaining two are the counterfactual event ($X_{CE}$) and the ending edited to reflect the counterfactual event ($Y_{EE}$). These notations are consistently used throughout the paper.

\begin{table}[h]
\small
  \centering
  \begin{tabular}{@{}ll@{}}
    \toprule
    \textbf{Field} & \textbf{Symbol} \\ \midrule
    Premise                     & $X_P$  \\
    Initial event               & $X_{IE}$ \\
    Original ending             & $X_{OE}$ \\
    Counterfactual event        & $X_{CE}$ \\
    Edited ending            & $Y_{EE}$ \\ \bottomrule
  \end{tabular}
  \caption{Structure of a story}
  \label{tab:sample_structure}
\end{table}

According to the definition of this task, the human annotators of $Y_{EE}$ were asked to incorporate the counterfactual event into the narrative with minimal modifications, while preserving overall coherence. An example of the various fields is provided in Table~\ref{tab:story_example}, where the differences between the initial and the counterfactual events (``seemed nice enough'' vs. ``was too rainy''), and those between the original and edited endings (e.g., ``sneezing'', ``allergies'' vs. ``soaked through'', ``wet'', ``cold''), are  highlighted. 
Given an input tuple $(X_P, X_{IE}, X_{OE}, X_{CE})$, the main inference objective is for a model to generate a predicted ending, $\hat{Y}_{EE}$, that closely aligns with the reference edited ending, $Y_{EE}$.

\begin{table}[h]
\small
\centering
\begin{tabular}{@{}p{2.5cm}p{4.9cm}@{}}
\toprule
\textbf{Field} & \textbf{Example} \\ \midrule
Premise & I tried going to the park the other day. \\[2pt]
\midrule
Initial event & The weather \textcolor{red}{seemed nice enough} for a walk. \\[2pt]
\midrule
Original ending &  Within minutes of getting there I \textcolor{blue}{started sneezing}. My \textcolor{blue}{eyes} were \textcolor{blue}{watery} and it was \textcolor{blue}{hard to breathe}. \textcolor{blue}{My allergies} were too \textcolor{blue}{bad} and I had to go back home. \\[2pt]
\midrule
Counterfactual event & The weather \textcolor{pink}{was too rainy} for a walk. \\[2pt]
\midrule
Edited ending & Within minutes of getting there I \textcolor{green}{was soaked through}. My \textcolor{green}{clothes} were \textcolor{green}{wet} and it was \textcolor{green}{cold}. My \textcolor{green}{clothes} were too \textcolor{green}{wet} and I had to go back home. \\
\bottomrule
\end{tabular}
\caption{A counterfactual story rewriting example. We use colors to highlight the main differences between, respectively, the initial event and the counterfactual event (\textcolor{red}{red} vs \textcolor{pink}{pink}) and those between the original ending and the edited ending (\textcolor{blue}{blue} vs \textcolor{green}{green}). Equivalent comments are provided in the accompanying text for color-blind readers.}
\label{tab:story_example}
\end{table}

\subsection{Counterfactual Rewriting Score}
\label{sec:CRM}

Counterfactual story rewriting is a natural language generation task that can be evaluated with any of the common metrics for these tasks, such as BLEU~\cite{bleu}, ROUGE~\cite{rouge} and many others (see Table \ref{tab:metrics}). In this study, we have decided to use both ROUGE and BLEU as traditional n-gram-based metrics alongside BARTScore~\cite{bartscore}, a more recent, model-based evaluation metric. Unlike n-gram-based metrics, BARTScore is based on a pre-trained language model that estimates the log-likelihood of a hypothesis given a source input, effectively capturing the semantic alignment between the two. 
Its tolerance to lexical variations makes it well suited for measuring meaning preservation, whereas both BLEU and ROUGE focus on the overlap of specific textual elements.
Together, these metrics provide complementary insights: BLEU and ROUGE confirm the inclusion of important content, while BARTScore assesses the semantic accuracy and fluency of the generated ending.

However, metrics such as BLEU, ROUGE and BARTScore simply compare model outputs against reference texts, and while simple to compute and well accepted, they are not fully aligned with the goals of counterfactual rewriting. In this task, the essential changes tend to be subtle and localized, while much of the surrounding narrative remains intact. As a consequence, metrics based on global similarity may overvalue unchanged segments and overlook the importance of the targeted counterfactual revisions. 

To avoid this shortcoming, we propose employing a specialized evaluation approach tailored to the nuances of counterfactual rewriting. Our key idea is to leverage a \emph{delta} ($\Delta$) score that explicitly quantifies the model’s ability to produce outputs that are more aligned with the revised, counterfactual version than with the original. Given a model output $\hat{Y}_{EE}$, the original ending $X_{OE}$, and the desired counterfactual ending $Y_{EE}$, the delta score is defined as:

\begin{equation}
\label{eq:DM}
\Delta_{M} = M(\hat{Y}_{EE}, Y_{EE}) - M(\hat{Y}_{EE}, X_{OE}),
\end{equation}

\noindent where $M$ can be any standard evaluation metric. The $\Delta_{M}$ score evaluates whether the generated output is closer to the edited ending that we target rather than the original ending; a higher value suggests a more successful incorporation of the counterfactual condition. It is, by design, a contrastive measure similar to those commonly used in contrastive pre-training \cite{CLIP} and preference optimization \cite{dpo2023, Xu2024CPO}.

However, since $\Delta_{M}$ is a relative score, it may fail to capture cases where the model performs poorly against both texts. To address this, we also report an adjusted score that incorporates both absolute and relative performance:

\begin{equation}
\begin{split}
\label{eq:score_delta}
& M(\hat{Y}_{EE}, Y_{EE}) + \Delta M = \\
& 2M(\hat{Y}_{EE}, Y_{EE}) - M(\hat{Y}_{EE}, X_{OE}).
\end{split}
\end{equation}

With this formulation, this adjusted score is able to simultaneously reward a) the matching with the edited ending (accuracy) and b) the matching improvement with respect to the original ending (relative improvement).

Our delta-based framework is general and can be implemented with any standard similarity measure. Moreover, it can be extended to include other forms of contrastive evaluation---such as, for instance, measuring the change in similarity with respect to the counterfactual event versus the initial event. Such dedicated measures can enable a more accurate and task-aligned assessment of a model's performance in counterfactual rewriting, offering a valuable alternative or complement to traditional sentence-level similarity metrics.

\subsection{The Proposed Differentiable Training Objective (DTO)}
\label{sec:dto}

Our goal is to employ the dedicated metrics in Equations \ref{eq:DM} and \ref{eq:score_delta} as the training objective for our model. However, two immediate difficulties arise: 1) conventional metrics such as BLEU and ROUGE are essentially counting measures, and, as such, are outright non-differentiable; 2) even differentiable models such as BARTScore receive in input a sequence of discrete tokens, and if we were to provide our model's prediction, $\hat{Y}_{EE}$, as the input, we would not be able to backpropagate the gradients through to our model due to the discreteness of the predicted tokens. 

For this reason, we choose to ``soften'' the output of our generative model by replacing it with differentiable embeddings, allowing us not to break the backpropagation chain. In detail, let us refer to BARTScore as the metric, and note the output probability of our generative model at slot $t$ as $p_t$. This is a vector of probability values over the generative model's vocabulary, $V$. Let us assume that the BARTScore model uses the same vocabulary, and that we have access to its input embedding matrix, $E$, of size $|V| \times D$, where $D$ is the size of an individual input embedding. We can therefore multiply $p_t$ by $E$, to obtain an ``expected'' embedding, $e_t$:

\begin{equation}
e_t = p_t^\top E
\label{eq:e_t}
\end{equation}

\noindent that can be used as input to BARTScore for the $t$-th slot of our prediction. Such an embedding does not correspond to any individual embedding in $E$, but, given that the embedding space is continuous, it will reflect the probability distribution in $p_t$ and capture its semantic, acting as a ``soft token'' or ``soft'' prediction. To train our model, we then compute the gradients of the log probability of the reference sentence in the BARTScore model, back-propagate them through the BARTScore model itself (frozen), and, eventually, through our generative model for its parameter update.

This simple scheme also allows for interesting variations: for instance, \cite{jauregi2021berttune} have proposed adding a step of Gumbel-softmax sampling \cite{Gumbel} using $p_t$ as the mean before the multiplication. This allows greater exploration of the output space, controllable via the Gumbel-softmax' temperature parameter, and for this reason we have adopted it in our work. In the more general case in which the vocabularies of the generative model and BARTScore differ, it is always possible to align them with simple heuristics, as done by \cite{TomLREC2024}.

\subsection{Loss Function}

The BARTScore score is simply computed by an adequately pre-trained BART model as the log probability of a given sentence hypothesis, $h$, for a given input text, $x$. The score can be expressed as:

\begin{equation}
\label{eq:BARTScore}
\textsc{BARTScore}(x,h)
=
\frac{1}{T}\sum_{t=1}^T
\log p\bigl(h_t \mid x,\,h_{<t}\bigr)
\end{equation}

In our case, we aim to use the scores in Equations \ref{eq:DM} and \ref{eq:score_delta} ($\Delta_M$, $M(\hat{Y}_{EE}, Y_{EE}) + \Delta_M$, and also $M(\hat{Y}_{EE}, Y_{EE})$ alone) as the training objective, using BARTScore as the base metric and replacing our prediction $\hat{Y}_{EE}$ with the soft prediction, that we note concisely as $E_\theta(\hat{Y}_{EE})$. For instance, in the case of the $M(\hat{Y}_{EE}, Y_{EE})$ score, our training objective becomes:

\begin{equation}
\label{eq:reward}
\begin{split}
& \textsc{BARTScore}(E_\theta(\hat{Y}_{EE}),Y_{EE})\\
& = \frac{1}{T}\sum_{t=1}^T
\log p\bigl(Y_{EEt} \mid E_\theta(\hat{Y}_{EE}),\,Y_{EE<t}\bigr)
\end{split}
\end{equation}

\noindent with obvious modifications for the other scores. Eventually, since the loss function is to be minimized by convention, we set:
\begin{equation}
L_{\mathrm{DTO}}(\theta)
=
-\textsc{BARTScore}(E_\theta(\hat{Y}_{EE}), Y_{EE}).   
\end{equation}

\vspace{1pt}

\subsection{Variance Reduction and Stability}

\begin{figure}[H]
    \centering

    \begin{subfigure}{\linewidth}
        \centering
        \includegraphics[width=0.9\linewidth, height=3cm]{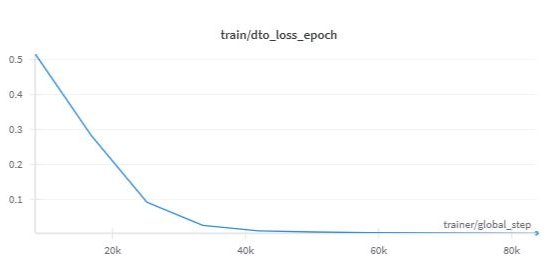}
        \caption{Training loss.}
        \label{fig:train}
    \end{subfigure}

    \vspace{1em} 

    \begin{subfigure}{\linewidth}
        \centering
        \includegraphics[width=0.9\linewidth, height=3cm]{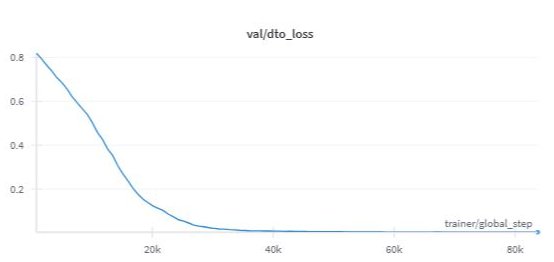}
        \caption{Validation loss.}
        \label{fig:valid}
    \end{subfigure}

    \caption{An example of the training loss for the score in Equation \ref{eq:DM} and the  corresponding validation loss.}
    \label{fig:losses}
\end{figure}

The classic approach for optimizing a non-differentiable evaluation metric is to employ the policy gradient \cite{williams1992simple}. However, most policy gradient approaches suffer from high variance due to the impact of sampling, which tends to affect convergence and stability. To mollify the variance, sophisticated baselines have been proposed (e.g., \cite{RELAX}), but in many cases, they introduce a bias in the gradient~\cite{SuttonB98}. By contrast, the DTO approach that we propose in our paper is intrinsically unbiased and is not affected by the variance issue since it does not rely on sampling. Furthermore, averaging over \(T\) tokens and \(N\) samples in each batch acts as an additional variance‐reduction mechanism, smoothing out the per‐token fluctuations in the log probabilities. To this aim, Figure \ref{fig:train} shows a plot of the training loss at epoch points for our delta score in Equation \ref{eq:DM}, displaying a very smooth and stably-converging behavior. In turn, Figure \ref{fig:valid} shows that the  corresponding validation loss follows a similar trend, giving empirical evidence to the efficacy of the proposed training approach.

\section{Experimental Set-Up}
\label{sec:experiments}

In this section we describe the data, model configurations, baselines, and evaluation protocols used to assess the effectiveness of our proposed DTO approach in counterfactual story rewriting.

\subsection{Datasets}
For our experiments, we have employed the \textsc{TimeTravel} \cite{qin2019counterfactual} and the Abductive Reasoning Task (\textsc{ART}) \cite{bhagavatula2020abductive} datasets. 
Both these datasets have been derived from the \textsc{ROCStories} corpus \cite{mostafazadeh2016corpus}, but they have been independently annotated for different tasks. In particular, the \textsc{TimeTravel} dataset has been annotated for counterfactual rewriting tasks, and for this reason each sample contains an original and (at least) a reference edited ending. The dataset's splits consist of a) a training set with 16,752 samples, each annotated with a single edited ending; and b) a validation and test sets both with 1,871 samples, each annotated with three, slightly different edited endings. Providing multiple references in the validation and test sets aims to make the evaluation more robust against ending variations. In our experiments, we have simply treated the different endings as separate samples.

At its turn, the ART dataset has been annotated for a main task known as abductive reasoning. In this task, each sample consists of a premise, an ending, and two alternative, intermediate events, only one of which is consistent with the ending while the other is in contradiction. All the fields are provided in input to a predictive model that has to decide which of the two intermediate events is consistent with the ending. The dataset contains 72,046 examples, divided into 50,481 training, 14,313 validation, and 7,252 test instances. Given that our focus is on  generative tasks, we reuse this dataset in a different way: we provide in input the premise  and the two intermediate events, and we predict the ending. This probes a model's ability to leverage the correct intermediate event without being distracted by the contradictory one, and acts as an ablative task compared to story rewriting where the correct ending has to be predicted without the provision of the original ending in input.

\begin{table}[h]
\centering
\caption{Main figures for the datasets}
\label{tab:dataset_composition}
\begin{tabular}{@{}lll@{}}
\toprule
Dataset & TimeTravel & ART \\
\midrule
\textsc{Training} & 16,752 & 50,481 \\
\textsc{Validation}\textsuperscript{*} & 5,613 & 14,313 \\
\textsc{Test}\textsuperscript{*} & 5,613 & 7,252 \\
\bottomrule
\end{tabular}
\begin{tablenotes}
\footnotesize
\item[*] *Each story includes three edited endings, labeled individually by crowd-sourced annotators. We treat each story ending as a separate instance.
\end{tablenotes}
\end{table}

Table~\ref{tab:dataset_composition} provides a summary of the two datasets. The \textsc{TimeTravel} dataset's edited endings consist on average of three sentences each, totaling 50 tokens on average. In contrast, the ART dataset's edited endings mainly consist of a single, brief sentence averaging 10 tokens. As such, \textsc{TimeTravel}'s endings are more complex than ART's.


\subsection{Main Settings}
\label{sec:settings}

As base model to implement our approach we have used \texttt{BART-large-cnn}~\cite{bart}, a transformer with 406M parameters originally specialized for summarization tasks. The main reason for selecting this model is that it is a language model of very small size (<1B parameters) that can be fine-tuned and operated with very limited computational costs, and has been specialized for fluent and articulate generation. We refer to this model as BART hereafter. 
To prepare the input for this model, we simply concatenate the various elements of each story, separating consecutive elements with a dedicated token (\texttt{</s>}) and making sure that the input does not exceed the maximum acceptable length (1024 tokens). Full training details for our model are provided in Appendix \ref{app:hyperparameters}.

In the experiments, we evaluate the model's predictions against the provided edited endings by using BARTScore, ROUGE-L, and SacreBLEU \cite{post-2018-call} as the metrics, using the two scores presented in Equations \ref{eq:DM} and \ref{eq:score_delta}. Collectively, these metrics are able to assess the fidelity, semantic and correct adaptability of the generated endings.

Our key performance comparison is between our models and two identical models trained, respectively, with the conventional NLL and CPO \cite{Xu2024CPO}. As reported in \cite{Xu2024CPO}, CPO is a high-performing and computationally-efficient variant of direct preference optimization. Its performance is influenced by two main hyperparameters: $\beta$, that sets the ``temperature'' of the preference-based objective, and $\lambda$, that balances the preference-based objective with the standard NLL (details in Appendix \ref{app:CPO}). In its implementation, we have introduced a small modification by computing the loss components at batch level rather than at sentence level, as it seemed to introduce more stability. In addition, we have also carried out experiments using two LLMs, namely GPT-4o (specifically, gpt-4o-2024-08-06\footnote{https://platform.openai.com/docs/models\#gpt-4o}), and Gemini 2.0 Flash (gemini-2.0-flash-001\footnote{https://cloud.google.com/vertex-ai/generative-ai/docs/gemini-v2\#2.0-flash}; simply Gemini 2.0 hereafter). GPT-4o's size is not officially disclosed, but it is estimated to be 200B parameters, while Gemini 2.0 is rumored to be in the vicinity of 70B\footnote{Source: https://x.com/ArtificialAnlys/status/1867292015181942970}. These models were used with four different prompting configurations: 1) zero-shot; 2) one-shot with a random example; 3) one-shot with a fixed example; and 4) one-shot with an example retrieved for the specific query with retrieval-augmented generation (RAG). All the configuration details are provided in Appendixes \ref{app:LLM} and \ref{app:RAG}. While other results on the \textsc{TimeTravel} dataset are available from the literature (e.g., \textsc{Delorean} \cite{qin-etal-2020-back}, \textsc{EduCat} \cite{chen2022educat}, SCM \cite{Wang2024SCM}), we believe that they have generally been superseded by these more contemporary baselines.

\section{Results}
\label{sec:results}

\begin{table*}[t]
\scriptsize
\begin{tabularx}{\textwidth}{l *{6}{>{\centering\arraybackslash}X}}
\toprule
& \multicolumn{2}{c}{BARTScore} & \multicolumn{2}{c}{ROUGE-L (\%)} & \multicolumn{2}{c}{SacreBLEU} \\
\cmidrule(lr){2-3} \cmidrule(lr){4-5} \cmidrule(lr){6-7}
\textbf{Method} & $\hat{Y}_{EE}, Y_{EE}$ & $\hat{Y}_{EE}, Y_{EE} + \Delta_{M}$ & $\hat{Y}_{EE}, Y_{EE}$ & $\hat{Y}_{EE}, Y_{EE} + \Delta_{M}$ & $\hat{Y}_{EE}, Y_{EE}$ & $\hat{Y}_{EE}, Y_{EE} + \Delta_{M}$ \\
\midrule
\textbf{NLL} & -1.710 & -2.885 & 60.9 & 45.3 & 34.5 & 18.0 \\
\midrule
\textbf{DTO (Ours)} & & & & & & \\
\midrule
DTO-Delta & -1.730 & -2.813 & 60.6 & 46.2 & 34.6 & 19.6 \\
DTO-Score & \textbf{-1.683} & -2.879 & \textbf{61.5} & 46.1 & \textbf{35.5} & 19.2 \\
DTO-Score+Delta  & -1.695 & -2.864 & 61.2 & 46.1 & 35.3 & 19.3 \\
\midrule
\textbf{CPO ($\beta$, $\lambda$)} & & & & & & \\
\midrule
CPO (0.1, 2)  &-1.709 &-2.835&60.2&45.9&34.8&19.5\\
CPO (1, 1) & -1.772 & -2.757 & 59.5 & \textbf{46.3} & 33.4 & 19.8\\
CPO (2, 0.5) & -1.787 & -2.762  & 58.9 & 45.7 & 32.8 & 19.6\\
\midrule
\textbf{GPT-4o} & & & & & & \\
\midrule
Zero-Shot & -2.290 & -2.873 & 46.9 & 38.9 & 7.4 & 9.2 \\
One-Shot Random & -2.480 & -2.788 & 41.4 & 37.6 & 8.2 & 8.3 \\
One-Shot Fixed & -2.872 & -2.981 & 28.4 & 27.9 & 8.2 & 8.3 \\
One-Shot RAG & -2.439 & -2.783 & 42.4 & 38.5 & 18.7 & 15.0 \\
\midrule
\textbf{Gemini 2.0} & & & & & & \\
\midrule
Zero-Shot & -3.081 & -3.210 & 22.2 & 21.4 & 7.3 & 9.8 \\
One-Shot Random & -2.712 & -2.967 & 35.8 & 32.2 & 8.5 & 8.8 \\
One-Shot Fixed & -2.537 & -2.745 & 40.5 & 37.2 & 25.8 & \textbf{31.4} \\
One-Shot RAG & -2.309 & \textbf{-2.690} & 50.7 & 43.9 & 31.0 & 24.0 \\
\bottomrule
\end{tabularx}
\caption{Model evaluation on the \textsc{TimeTravel} dataset using BARTScore, ROUGE-L, and SacreBLEU. The performance differences between our DTO-Score and DTO-Score+Delta models versus the NLL have been tested with a one-tailed nonparametric bootstrap test \cite{dror-etal-2018-hitchhikers} yielding a $p$-value of 0.}
\label{tab:merged_pg_NLL_ppo_results}
\end{table*}

Table \ref{tab:merged_pg_NLL_ppo_results} presents the results of our main experiment.
For each model, we report both the usual predictive score against the reference (noted concisely as $\hat{Y}_{EE}, Y_{EE}$) and our counterfactual rewriting score (noted as $\hat{Y}_{EE}, Y_{EE} + \Delta_{M}$). 
For our model, we present three different versions trained with respective variants of the training objective: $\Delta_M$ (\textit{DTO-Delta}); $\hat{Y}_{EE}, Y_{EE}$ (\textit{DTO-Score}); and $\hat{Y}_{EE}, Y_{EE} + \Delta_{M}$ (\textit{DTO-Score+Delta}).
For CPO, we report only the three best-performing configurations across the hyperparameter range reported in Appendix \ref{app:hyperparameters}.

As a preliminary comment, all the counterfactual scores have been lower than the corresponding predictive scores for all models and metrics. This is due to the fact that all the models \textit{receive in input the original ending}, and their predictions are therefore able to reflect it better than they can reflect the unseen, edited ending (i.e., $\hat{Y}_{EE}, X_{OE} > \hat{Y}_{EE}, Y_{EE}$ in general), leading to negative $\Delta_M$ scores as a consequence. With that said, the key comparative observation is that our DTO-Score and DTO-Score+Delta models have comprehensively outperformed the NLL model in both the predictive and counterfactual scores for all metrics. DTO-Score has achieved the best results in predictive score, while DTO-Score+Delta has achieved the best results in counterfactual score. Given that only BARTScore was used in the training objective, the substantial improvements also in ROUGE-L and SacreBLEU confirm that our models generalize well and avoid outright metric circularity. These results are remarkable as they show that the proposed approach has been able to improve the prediction of the edited ending compared to the NLL, while at the same time increasing the counterfactual gain with respect to the original ending. In contrast, the DTO-Delta model has delivered mixed results, with modest improvements in counterfactual score and a deterioration in predictive score, proving the importance of the compensation term introduced in our Equation \ref{eq:score_delta}.

Interestingly, our models have also outperformed the CPO models in almost all the metrics. Only the CPO (1,1) configuration has achieved a higher counterfactual BARTScore than our best model (i.e., -2.757 vs -2.879). In all the other metrics, our models have achieved relative improvements over the best CPO results (e.g., 61.5 predictive ROUGE-L score for our DTO-Score model vs 60.2 for CPO (0.1, 2)). It is also important to note that we have deliberately chosen the best-performing CPO configurations based on their results on the test set in order to make the comparison more challenging for our models.

The results for the two LLMs are also interesting: GPT-4o has performed rather well at zero-shot, yet without nearing the results of the NLL, DTO and CPO fine-tuned models. However, the one-shot configurations have showed a deterioration in most of the metrics, with the notable exception of the SacreBLEU score for the RAG prompting. In contrast, Gemini 2.0 has reported modest results at zero-shot, but improved remarkably in the one-shot configurations, reaching the best counterfactual scores of all in BARTScore and SacreBLEU. However, the respective predictive scores are much lower than those of the fine-tuned models, suggesting that the good counterfactual performance is achieved at the expense of the adherence to the edited ending. These results are even more impressive considering that the proposed model is only $\sim$0.4B parameters in size vs GPT-4o's 200B and Gemini 2.0's 70B (estimated). While the performance of the LLMs could certainly be improved by employing more aggressive prompting techniques (e.g., n-shots, chain-of-thought, step-back) or more recent models (GPT-5, Gemini 3), these results show that a small language model fine-tuned with an effective training objective can deliver a valuable performance with a fraction of their computational footprint.\\


    
    



\begin{table*}[t]
\centering
\fontsize{8.5}{8.8}\selectfont
\begin{tabular}{p{7.42cm} p{7.42cm}}
\toprule
\multicolumn{2}{l}{\textbf{Premise:} The soccer game was tied 3 to 3 and there was a minute left to play.} \\
\midrule
\textbf{Initial event:} Julie \textcolor{red}{had never scored a goal yet, but knew today would be her day}. & \textbf{Counterfactual event:} Julie \textcolor{pink}{was eagerly watching the game in the stands}.\\
\midrule
\textbf{Original ending:} Ashley passed \textcolor{blue}{her} the ball and this was chance. \textcolor{blue}{She} kicked as hard as she could, and the ball soared into the net. Julie's \textcolor{blue}{first} goal \textcolor{blue}{won} the game.
& \textbf{Edited ending:} Ashley passed the ball and this was the chance. \textcolor{green}{Another teammate} kicked as hard as she could, and the ball soared into the net. Julie's  \textcolor{green}{got to see the} goal \textcolor{green}{win} the game. \\
\bottomrule
\end{tabular}


\begin{tabular}{p{7.42cm} p{7.42cm}}
\toprule
\textbf{Words unique to the original ending (blue):} & \textbf{Words unique to the edited ending (green):} \\
\textcolor{blue}{her, she, first, won} & \textcolor{green}{another, teammate, got, to, see, the, win} \\
\midrule
\textbf{Words unique to the counterfactual event (pink):} & \textbf{Shared words/phrases (black):} \\
\textcolor{pink}{eagerly, watching, stand} & Ashley, passed, the, ball, and, this, was, chance,  kicked, as, hard, she, could, soared, into, net, Julie, is, goal,  game. \\
\bottomrule
\end{tabular}



\begin{tabular}{p{8cm} p{2cm} p{2cm} p{2cm}}
\toprule
\textbf{Predicted endings:} & \textbf{BARTScore} & \textbf{ROUGE-L} & \textbf{SacreBLEU} \\
\midrule
\textbf{DTO-Score+Delta:} Ashley had the ball and this was the team’s chance. \textcolor{blue}{She} kicked as hard as \textcolor{blue}{she} could, and the ball soared into the net. Julie cheered from the \textcolor{pink}{stands} as \textcolor{blue}{her} \textcolor{green}{team} \textcolor{blue}{won} the game. & -2.377 & 62.74 & 53.45 \\
\midrule
\textbf{GPT-4o Zero-Shot:} Ashley had the ball and this was their chance. \textcolor{blue}{She} kicked as hard as she could, and the ball soared into the net. Julie’s team \textcolor{blue}{won} the \textcolor{blue}{game} as \textcolor{blue}{she} cheered from the \textcolor{pink}{stands}, ecstatic for her \textcolor{green}{teammates}’ victory & -2.161 & 58.18 & 44.53 \\
\midrule
\textbf{GPT-4o One-Shot Fixed:} With the game tied 3 to 3 and only a minute left to play, Julie watched anxiously from the \textcolor{pink}{stands}, her heart pounding with every passing second. \textcolor{blue}{Her} \textcolor{green}{teammate} Ashley, always quick on her feet, intercepted the ball and expertly maneuvered & -3.487 & 16.67 & 4.63\\
\midrule
\textbf{GPT-4o One-Shot RAG:} Ashley passed her the ball and this was her chance. \textcolor{blue}{She} kicked as hard as \textcolor{blue}{she} could, and the ball soared \underline{just wide of the net}. The final whistle blew, and \underline{the game ended in a tie}. & -1.926
 & 66.66 & 43.61\\

\bottomrule
\end{tabular}


\caption{
Qualitative example - “soccer game's final minutes”. 
We use colors to highlight the main differences between the initial and counterfactual events (\textcolor{red}{red} vs \textcolor{pink}{pink}) and between the original and edited endings are (\textcolor{blue}{blue} vs \textcolor{green}{green}). Color-blind readers can find corresponding comments in the accompanying text.}
\label{tab:full_example}
\end{table*}

\paragraph{Qualitative Example} To better illustrate the behavior of the various models\footnote{For reasons of space, we only include our DTO-Score+Delta model and GPT-4o zero-shot, one-shot fixed and one-shot RAG. Gemini 2.0 has performed mildly worse than GPT-4o in all metrics for this example.}, we present a commented example in Table~\ref{tab:full_example}. This example requires transforming a story where Julie actively scores a goal (the original ending) into one where she is merely a spectator (the counterfactual event). Successful adaptation demands both \textit{removal} of agency-specific references (e.g., “her,” “she kicked,” “first goal”) and \textit{generation} of contextually appropriate substitutes that reflect Julie’s passive role. As the table shows, the prediction by our DTO model is rather accurate and introduces minimal edits compared to the original ending. While it attributes the goal to Ashley rather than another teammate, the counterfactual sense of the story is captured. In turn, the prediction by GPT-4o zero-shot is similar, with only a marked departure from the original and expected endings toward the end of the prediction.
However, the predictions by both one-shot variants are affected by major departures from the task's instructions. GPT-4o one-shot fixed attempted a longer prediction, leading to text truncation. While this could certainly be amended by allowing longer responses, it would violate the task's  requirement of minimal edits. At its turn, GPT-4o one-shot RAG predicted an outright wrong ending (a tie rather than a win), despite its higher BARTScore and ROUGE-L metrics.
The other observation is that all the responses by the GPT-4o variants tend to ``embellish" the predicted ending with elements that are not present in the original ending or the counterfactual event.


\paragraph{Ablated Task Analysis}

The \textsc{TimeTravel} dataset is the only public dataset expressly annotated for the task of counterfactual story rewriting. For this reason, we extend the evaluation of the proposed approach to an ``ablated'' version of this task, where the same fields are passed in input to the model, with the exception of the original ending. The aim of this task is to evaluate the model's ability to predict an ending based on the premise and the correct intermediate event, but without being ``distracted'' by the contradicting event.
For this evaluation, we employ both the ART dataset and a version of the \textsc{TimeTravel} dataset without the original ending in input (Ablated \textsc{TimeTravel}). For brevity, we limit the analysis to the NLL baseline and our model with the best predictive score, DTO-Score. Note that we cannot compute counterfactual scores for this task since $X_{OE}$ is not available.

\begin{table}[H]
\caption{Model evaluation on the ablated task.} 
\scriptsize
\label{tab:ablated_dto_mle_results}
\begin{tabular*}{\columnwidth}{@{\extracolsep\fill}lccc}
\toprule
\textbf{Method} 
    & \shortstack{BARTScore \\ $(\hat{Y}_{EE}, Y_{EE})$}
    & \shortstack{ROUGE-L \\ $(\hat{Y}_{EE}, Y_{EE})$}
    & \shortstack{SacreBLEU \\ $(\hat{Y}_{EE}, Y_{EE})$} \\
\midrule
\multicolumn{4}{l}{\textbf{Ablated \textsc{TimeTable} Dataset}} \\
\midrule
NLL & -2.295& 25.1 & 30.3\\
DTO-Score & \textbf{-2.082} & \textbf{26.0}& \textbf{30.6}\\
\addlinespace[2pt]
\midrule
\multicolumn{4}{l}{\textbf{ART Dataset}} \\
\midrule
NLL & -3.235 & \textbf{24.8} & 27.6 \\
DTO-Score & \textbf{-3.069} & 23.1 & \textbf{28.3} \\
\addlinespace[2pt]
\bottomrule
\end{tabular*}
\end{table}
 
Table \ref{tab:ablated_dto_mle_results} reports the results for this experiment, showing that the proposed DTO-Score approach has been able to outperform the NLL baseline in 5 scores out of 6, confirming its strong performance also on this task. As a side note, all these scores are lower than the corresponding scores for the non-ablated task because predicting the edited ending without the original ending in input is significantly more challenging.
 
\section{Conclusion}
\label{sec:Conclusion}

This paper has proposed a novel, differentiable training objective---DTO---for the task of counterfactual story rewriting. The key aims of the proposed objective are twofold: 1) training the generative model with an objective closer to the evaluation metrics for this task than the standard negative log-likelihood, in the hope of obtaining better performance at test time; and 2) circumventing the well-known limitations of non-differentiable training approaches such as the policy gradient in variance and stability. For this reason, we have proposed a counterfactual training reward based on a differentiable model (BARTScore), and ensured backpropagation to our generative model thanks to a mechanism of ``soft'' predictions. As such, the proposed approach is ensured to be end-to-end differentiable, avoiding the need for sampling and exhibiting stable convergence. The experimental results over the counterfactual \textsc{TimeTravel} dataset have shown that a BART model trained with our proposed DTO training objective has been able to outperform two identical models trained, respectively, with the conventional negative log-likelihood and contrastive preference optimization (CPO) in terms of predictive and counterfactual scores, and also perform competitively against two contemporary LLMs. While the LLM performance could undoubtedly be improved by employing more advanced models and prompting techniques, our results show that a small language model (0.4B parameters) with an effective training objective such as the one we have proposed can perform on this task comparably to commercial-grade models that are 100-500$\times$ larger. For this reason, in the future, we plan to expand our exploration of training objectives for small language models to other preference and contrastive differentiable training objectives and their potential combination.






\bibliographystyle{ieeetr}  
\bibliography{custom_dedup}

\vspace{12pt}

\appendices

\section{Direct and Contrastive Preference Optimization}
\label{app:CPO}

Rafailov \textit{et al}. in \cite{dpo2023} reformulated reinforcement learning from human feedback (RLHF) as a pairwise differentiable training objective. Given an input, $x$, and two candidate outputs, a \emph{preferred} output $y_w$ and a \emph{dispreferred} output $y_l$ (informally referred to as ``winner'' and ``loser'', hence the indexes), they defined the direct preference optimization (DPO) loss as:

\begin{equation}
\begin{split}
L(\pi_\theta;\pi_{\mathrm{ref}}) 
&= -\mathbb{E}_{(x,y_w,y_l)\sim\mathcal{D}} \Bigl[
   \log\sigma\Bigl(
   \beta \log \tfrac{\pi_\theta(y_w\mid x)}{\pi_{\mathrm{ref}}(y_w\mid x)} \\
&\quad\quad\quad\quad\quad\quad
 - \beta \log \tfrac{\pi_\theta(y_l\mid x)}{\pi_{\mathrm{ref}}(y_l\mid x)}
   \Bigr)
   \Bigr]
\end{split}
 \label{eq:dpo}
\end{equation}

Here, $\pi_\theta$ is the trainable policy model, $\pi_{\mathrm{ref}}$ is a fixed reference language model, $\beta$ is an inverse temperature parameter controlling the sharpness of preferences and the adherence of $\pi_\theta$ to $\pi_{\mathrm{ref}}$, and $\sigma(\cdot)$ denotes the sigmoid function. This objective encourages $\pi_\theta$ to increase the likelihood of preferred outputs ($y_w$) and suppress the likelihood of dispreferred outputs ($y_l$), relative to the baseline set by $\pi_{\mathrm{ref}}$. The DPO formulation introduces a contrastive signal inspired by the proximal policy optimization (PPO) framework \cite{SchulmanWDRK17}, but avoids reinforcement learning by directly optimizing this surrogate loss.

In order to increase the computational efficiency of the preference optimization, \cite{Xu2024CPO} simplified the DPO objective by removing the reference model, implicitly replacing $\pi_{\mathrm{ref}}$ with a uniform distribution $U(\cdot)$ and yielding the following simplified objective:

\begin{equation}
\begin{split}
L(\pi_\theta;U) 
&= -\mathbb{E}_{(x,y_w,y_l)} \Bigl[
     \log\sigma\bigl(
     \beta \log \pi_\theta(y_w\mid x)\\
&\quad
     - \beta \log \pi_\theta(y_l\mid x)
     \bigr)
   \Bigr]
\end{split}
\label{eq:cpo}
\end{equation}


In addition, to prevent the model from improving the contrastive objective at the expense of the likelihood of $y_w$, they added it back to the objective as a regularization term:

\begin{equation}
\begin{split}
L(\pi_\theta;U) 
&= -\mathbb{E}_{(x,y_w,y_l)} \Bigl[
     \log\sigma\bigl(
     \beta \log \pi_\theta(y_w\mid x)\\
&\quad
     - \beta \log \pi_\theta(y_l\mid x)
     \bigr) + \lambda \log \pi_\theta(y_w\mid x)
   \Bigr]
\end{split}
\label{eq:cpo_full}
\end{equation}

\noindent where we have introduced an extra hyperparameter, $\lambda$, to control the intensity of this regularization. This contrastive preference optimization (CPO) formulation retains the essential contrastive learning signal of DPO, but significantly reduces the computational cost by eliminating the need for the storage of and a forward pass through $\pi_{\mathrm{ref}}$. 

\medskip
Although not derived from it, the analogies between our proposed DTO approach and CPO are remarkable. In our $\Delta_M$ metric (Equation \ref{eq:DM}), the \emph{edited endings} serve as the preferred outputs ($y_w$), and the corresponding \emph{original endings} are treated as dispreferred outputs ($y_l$). As such, our using $\Delta_M$ as the training objective is analogous to using CPO's contrastive objective in Equation \ref{eq:cpo}. In addition, our using $M(\hat{Y}_{EE}, Y_{EE}) + \Delta_M$ as the training objective is analogous to using CPO's regularized contrastive objective in Equation \ref{eq:cpo_full}. The key difference between our proposed objectives and CPO's is that we directly optimize a desirable evaluation metric rather than a generic likelihood, increasing the chances that our trained model will perform along the expectations of the targeted task.

\section{Model Hyperparameters}
\label{app:hyperparameters}

All our experiments have used the pretrained \texttt{facebook/bart-large-cnn} model as the base model. This model was fine-tuned for the counterfactual rewriting task on a single NVIDIA A40 GPU with 48 GB of RAM using the PyTorch Lightning library. All the main hyperparameters and experimental settings are reported in Table~\ref{tab:model_general_hyperparameters} to facilitate reproducibility.

\begin{table}[t]
\caption{Model hyperparameters and settings}
\label{tab:model_general_hyperparameters}
\centering
\small
\begin{tabular}{@{}ll@{}}
\toprule
\textbf{Hyperparameter} & \textbf{Value} \\
\midrule
Model Name & BART-large-cnn \\
GPU & A40 (48 GB) \\
Learning Rate & 5e-9 \\
Max Input Length & 1024 \\
Max Output Length & 250 \\
Batch Size & 2 \\
DTO Epochs & 10 \\
Num Workers & 3 \\
Attention Output & Enabled \\
Gumbel-Softmax & Yes (T = 1.0, Hard = False) \\
Annealing & Disabled \\
CPO: $\beta$ & 0.1, 0.5, 1.0, 2.0 \\
CPO: $\lambda$  & 0.5, 1.0, 2.0 \\
\midrule
\multicolumn{2}{@{}l}{\textbf{Evaluation Metrics}} \\
BARTScore & Enabled \\
BLEU, ROUGE-L & Enabled \\
BERTScore Model & DeBERTa-MNLI \\
BERT Batch Size & 4 \\
Scorer Device & cuda:0 \\
\bottomrule
\end{tabular}
\end{table}

\section{LLM Configuration}
\label{app:LLM}

The two LLMs have been evaluated in four configurations:
\begin{itemize}
    \item \textbf{Zero-shot inference}: The model generates outputs without any demonstration examples, testing its ability to generalize from the task description alone.
    \item \textbf{Random one-shot inference}: The model is provided with a randomly selected example along with the prompt, which tests the model’s adaptability to leverage diverse instances as examples.
    \item \textbf{Fixed one-shot inference}: The model is given a fixed example for all the test samples to encourage consistency in the inference process.
    \item \textbf{Retrieval-augmented generation (RAG)}: For each sample in the test set, the model is given an example retrieved from the training set based on closest embedding similarity.
\end{itemize}

We acknowledge that other configurations ($n$-shot, chain-of-thought etc) may be able to achieve even better performance. However, these LLMs are already much larger than our BART model and we believe that they suffice as baselines. 

To determine an appropriate token limit for the generation by the LLMs, we have conducted a preliminary analysis of the length of the edited endings. The results are as follows:

\begin{itemize}
    \item Mean (\(\mu\)): 140.93 characters
    \item Standard deviation (\(\sigma\)): 29.94 characters
    \item Standard number of characters: \(\mu + 2\sigma = 140.93 + 2 \times 29.94 = 200.81\)
\end{itemize}

According to \cite{tactiq2024}, a token is, on average, approximately 4 characters. Therefore, the corresponding standard number of tokens is $200.81/4 \approx 50 \text{ tokens}$, which we assume to be a desirable token limit for the LLMs.

\vspace{6pt}

\textit{LLM prompt}. As prompt for the LLMs, we have used the following:

\vspace{6pt}

\begin{adjustwidth}{0pt}{0pt}
\texttt{Generate the adapted ending to fill these three aspects:
\begin{enumerate}
    \item Minimal Intervention: Adjust the story's original ending with minimal changes needed to align it with the counterfactual event. The edited ending should remain as close as possible to the original ending.
    \item Narrative Insight: Understand the story structure and make changes essential for maintaining the story's coherence and thematic consistency, avoiding unnecessary alterations.
    \item Counterfactual Adaptability: Adapt the story's course in response to the counterfactual event that diverges from the initial event.
\end{enumerate}
Premise: \{test\_data['premise']\} \\
Initial event: \{test\_data['initial']\} \\
Original ending: \{test\_data['original\_ending']\} \\
Counterfactual event: \\\{test\_data['counterfactual']\} \\\\
Now, generate the adapted ending:}
\end{adjustwidth}

\vspace{6pt}

This prompt is designed to ensure that the model generates an ending that adheres to the three critical aspects of counterfactual story rewriting: minimal intervention, narrative insight, and counterfactual adaptability, similarly to the instructions provided to the annotators.

\section{RAG Implementation Details}
\label{app:RAG}
Our RAG pipeline combines retrieval and generation components through a modular design. The implementation uses ChromaDB as the vector store for its balance of persistence and metadata support. The documents contain the story's text and structured metadata (story IDs, fields, event tags) to enable hybrid retrieval. We compared OpenAI's text-embedding-3-large (3072-dimensional) and Gemini's text-embedding-004 (768-dimensional) embeddings.

\begin{table}[h]
\caption{Main details of the RAG implementation}
\label{tab:rag_comparison}
\centering
\fontsize{8}{8}\selectfont
\begin{tabular}{@{}lll@{}}
\toprule
\textbf{Component} & \textbf{GPT-4o} & \textbf{Gemini 2.0} \\
\midrule
Embedding Model & text-embedding-3-large & text-embedding-004 \\
Dimensions & 3072 & 768 \\
Retrieval & k-NN (cos $\theta$) & k-NN (cos $\theta$) \\
Max Context & 128K tokens & 256K tokens \\
Throughput & 120 tok/s & 400 tok/s \\
Determinism & temp = 0.0 & temp = 0.0 \\
Storage/10K & 2.1 GB & 0.5 GB \\
Latency & 1.8 s & 0.9 s \\
\bottomrule
\end{tabular}
\end{table}

In terms of memory footprint, the GPT-4o implementation required 2.1 GB of storage for 10,000 stories, while Gemini 2.0 achieved a comparable functionality with only 0.5 GB thanks to its lower-dimensional embeddings. Retrieval operations had consistent $\sim$50 ms query latency across both systems for datasets of this scale. In end-to-end generation tasks, GPT-4o averaged 1.8 seconds per query compared to Gemini's 0.9 seconds, reflecting the latter's throughput optimizations.

The implementation prioritized reproducibility through ChromaDB's persistent storage backend. Identical metadata schemas were maintained across both GPT and Gemini to ensure fair comparison. Similarity scores were normalized to compensate for dimensional differences between embedding spaces. Both systems enforced deterministic outputs through temperature parameter locking (0.0) during generation.

\end{document}